\documentclass[letterpaper, 10 pt, conference]{ieeeconf}
\IEEEoverridecommandlockouts 
\usepackage{graphicx}
\usepackage[normalem]{ulem}
\usepackage[flushleft]{threeparttable}
\usepackage{amsmath,amsfonts,amssymb}
\usepackage{algorithmic}
\usepackage{algorithm}
\usepackage{array}
\usepackage[caption=false,font=footnotesize]{subfig}
\usepackage{textcomp}
\usepackage{stfloats}
\usepackage{url}
\usepackage{verbatim}
\usepackage{siunitx}
\usepackage{tabularx}
\usepackage{relsize}
\usepackage{xcolor}
\usepackage{textcase}

\usepackage[backend=bibtex,bibstyle=ieee,citestyle=numeric-comp]{biblatex}
\title{\LARGE \bf Optimal Robot Formations: Balancing Range-Based Observability and User-Defined Configurations}

\author{Syed Shabbir Ahmed, Mohammed Ayman Shalaby, Jerome Le Ny, and James Richard Forbes%
\thanks{This work was supported by the NSERC Discovery and Alliance Grant programs, and the Canadian Foundation for Innovation (CFI) program.}
\thanks{S. S. Ahmed, M. A. Shalaby, and J. R. Forbes are with the Department of
Mechanical Engineering, McGill University, 817 Sherbrooke St. W., Montreal, QC
H3A 0C3, Canada. J. Le Ny is with the Department of Electrical Engineering, Polytechnique Montreal, Montreal, QC H3T 1J4, Canada. \{{\tt\smaller syed.shabbir.ahmed@mail.mcgill.ca\}}.}%
}

\newcommand{\ignore}[1]{}  

\newcommand{\mc}[1]{\ensuremath{\mathcal{#1}}}

\newcommand{\mbc}[1]{\ensuremath{\boldsymbol{\mathcal{#1}}}}
\DeclareMathAlphabet{\mbf}{OT1}{ptm}{b}{n}

\newcommand{\mbfdot}[1]{\ensuremath{\dot{\mbf{#1}}}}

\newcommand{\mbfbar}[1]{\ensuremath{\bar{\mbf{#1}}}}
\newcommand{\mbfhat}[1]{\ensuremath{\hat{\mbf{#1}}}}
\newcommand{\mbfcheck}[1]{\ensuremath{\check{\mbf{#1}}}}

\newcommand{\mbfdel}[1]{\ensuremath{\delta{\mbf{#1}}}}
\newcommand{\mbftilde}[1]{\ensuremath{\tilde{\mbf{#1}}}}

\newcommand{\norm}[1]{\ensuremath{\left\Vert#1\right\Vert}}

\newcommand{\trans}{{\ensuremath{\mathsf{T}}}}

\newcommand{\trace}{{\ensuremath{\mathrm{tr}}}}

\newcommand{\diag}{{\ensuremath{\mathrm{diag}}}}

\newcommand{\utimes}{{\raisebox{-0.6ex}{\kern-1.0ex\raisebox{0.6ex}{\small$\mathsf{v}$}}}}

\newcommand{\rnums}{\mathbb{R}}

\DeclareMathOperator*{\argmin}{arg\,min}

\newcommand{\bma}[1]{\left[\begin{array}{#1}}  
\newcommand{\ema}{\end{array}\right]}
\newcommand{\bbm}{\begin{bmatrix}}  
\newcommand{\ebm}{\end{bmatrix}}
\newcommand{\beq}{\begin{equation}}  
\newcommand{\eeq}{\end{equation}}
\newcommand{\bdis}{\begin{displaymath}}  
\newcommand{\edis}{\end{displaymath}}
\newcommand{\beqarray}{\begin{eqnarray}}
\newcommand{\eeqarray}{\end{eqnarray}}
\newcommand{\beqarraynn}{\begin{eqnarray*}}
\newcommand{\eeqarraynn}{\end{eqnarray*}}

\newcommand{\SO}[1]{\ifmmode SO(#1)\else $SO(#1)$\fi}
\newcommand{\SE}[1]{\ifmmode SE(#1)\else $SE(#1)$\fi}

\newcommand{\so}[1]{\ifmmode \mathfrak{so}(#1)\else $\mathfrak{so}(#1)$\fi}
\newcommand{\se}[1]{\ifmmode \mathfrak{se}(#1)\else $\mathfrak{se}(#1)$\fi}

\newcommand{\dcmspace}{\hspace{0.1em}}
\NewDocumentCommand{\dcm}{ O{} }{\mbf{C}_{#1} {\ifthenelse{\equal{#1}{}}{}{\dcmspace}}}
\NewDocumentCommand{\dcmbar}{ O{} }{\mbfbar{C}_{#1} {\ifthenelse{\equal{#1}{}}{}{\dcmspace}}}
\NewDocumentCommand{\dcmcheck}{ O{} }{\mbfcheck{C}_{#1} {\ifthenelse{\equal{#1}{}}{}{\dcmspace}}}
\NewDocumentCommand{\dcmhat}{ O{} }{\mbfhat{C}_{#1} {\ifthenelse{\equal{#1}{}}{}{\dcmspace}}}
\NewDocumentCommand{\dcmdot}{ O{} }{\mbfdot{C}_{#1} {\ifthenelse{\equal{#1}{}}{}{\dcmspace}}}

\newcommand{\posespace}{\,}
\NewDocumentCommand{\pose}{ O{} O{} }{\mbf{T}_{#2}^{\ifthenelse{\equal{#1}{}}{}{\posespace} #1 \ifthenelse{\equal{#1}{}}{}{\posespace}}}
\NewDocumentCommand{\posebar}{ O{} O{}}{\mbfbar{T}_{#2}^{\ifthenelse{\equal{#1}{}}{}{\posespace} #1 \ifthenelse{\equal{#1}{}}{}{\posespace}}}
\NewDocumentCommand{\posecheck}{ O{} O{} }{\mbfcheck{T}_{#2}^{\ifthenelse{\equal{#1}{}}{}{\posespace} #1 \ifthenelse{\equal{#1}{}}{}{\posespace}}}
\NewDocumentCommand{\posehat}{ O{} O{} }{\mbfhat{T}_{#2}^{\ifthenelse{\equal{#1}{}}{}{\posespace} #1 \ifthenelse{\equal{#1}{}}{}{\posespace}}}
\NewDocumentCommand{\posedot}{ O{} O{} }{\mbfdot{T}_{#2}^{\ifthenelse{\equal{#1}{}}{}{\posespace} #1 \ifthenelse{\equal{#1}{}}{}{\posespace}}}

\begin{document}

%
%
%
%
%
%
%
\def \myJournal {IEEE/RSJ International Conference on Intelligent
Robots and Systems}
\def \myDoi {10.1109/IROS58592.2024.10801342}
\def \myPaperSiteName {IEEE Xplore}
\def \myPaperSiteLink {https://ieeexplore.ieee.org/document/10801342}
\def \myYear {2024}
\def \myPaperCitation{S. S. Ahmed, M. Shalaby, J. Le Ny and J. R. Forbes, ``Optimal Robot Formations: Balancing Range-Based Observability and User-Defined Configurations,'' in \textit{IEEE/RSJ International Conference on Intelligent Robots and Systems}, October 2024.}


\begin{figure*}[t]

\thispagestyle{empty}
\begin{center}
\begin{minipage}{6in}
\centering
This paper has been accepted for publication in \emph{\myJournal}. 
\vspace{1em}

This is the author's version of an article that has, or will be, published in this journal or conference. Changes were, or will be, made to this version by the publisher prior to publication.
\vspace{2em}

\begin{tabular}{rl}
DOI: & \myDoi\\
\myPaperSiteName: & \texttt{\myPaperSiteLink}
\end{tabular}

\vspace{2em}
Please cite this paper as:

\myPaperCitation

\vspace{15cm}
\copyright \myYear \hspace{4pt}IEEE. Personal use of this material is permitted. Permission from IEEE must be obtained for all other uses, in any current or future media, including reprinting/republishing this material for advertising or promotional purposes, creating new collective works, for resale or redistribution to servers or lists, or reuse of any copyrighted component of this work in other works.

\end{minipage}
\end{center}
\end{figure*}
\newpage
\clearpage
\pagenumbering{arabic} 

\maketitle
\thispagestyle{empty}
\pagestyle{empty}

\begin{abstract}
This paper introduces a set of customizable and novel cost functions that enable the user to easily specify desirable robot formations, such as a ``high-coverage'' infrastructure-inspection formation, while maintaining high relative pose estimation accuracy. The overall cost function balances the need for the robots to be close together for good ranging-based relative localization accuracy and the need for the robots to achieve specific tasks, such as minimizing the time taken to inspect a given area. The formations found by minimizing the aggregated cost function are evaluated in a coverage path planning task in simulation and experiment, where the robots localize themselves and unknown landmarks using a simultaneous localization and mapping algorithm based on the extended Kalman filter. Compared to an optimal formation that maximizes ranging-based relative localization accuracy, these formations significantly reduce the time to cover a given area with minimal impact on relative pose estimation accuracy. \color{black}
\end{abstract}

\section{Introduction}
The relative position and attitude between two robots, referred to as \emph{relative pose}, must be reliably estimated when conducting multi-robot tasks. Accurate relative pose estimation is essential for tasks such as collaborative planning and mapping, formation control, and coverage path planning. Cameras with object-detection ability or LiDAR in combination with other sensors can estimate the relative pose to within an acceptable accuracy \cite{Li2022Localization, SHEN2022, Zheng2023, Queralta2020, Xu2021, Guo2020, Nguyen2023}. However, the need for the robots to be in the cameras' field-of-view, the high cost and weight of LiDAR, as well as the substantial computational power required by both, hinder their use in many applications.

Recently, ultra-wideband (UWB) transceivers, referred to as UWB \emph{tags}, have been an increasingly popular choice for relative pose estimation due to their low cost, low weight, and low power consumption \cite{Yanjun2020SingleUwb, Hepp2016PersonTO, Samet2018, Sahinoglu2008}. The typical ranging accuracy for standard UWB tags is $10\,\si{cm}$ between a pair of transceivers. UWB tags are oftentimes fixed to static anchors with known locations and are then used to localize tags placed on mobile robots \cite{Moron2022, Shule2020UWB, Muller2015, Fang2021, Shen2010, Guo2017}.

\begin{figure}[t]
    \vspace{-1.5cm}
    \centering
    \subfloat[\label{fig:1a}]{\includegraphics[height = 6cm, trim={0cm 0.cm 0cm 0cm}, clip]{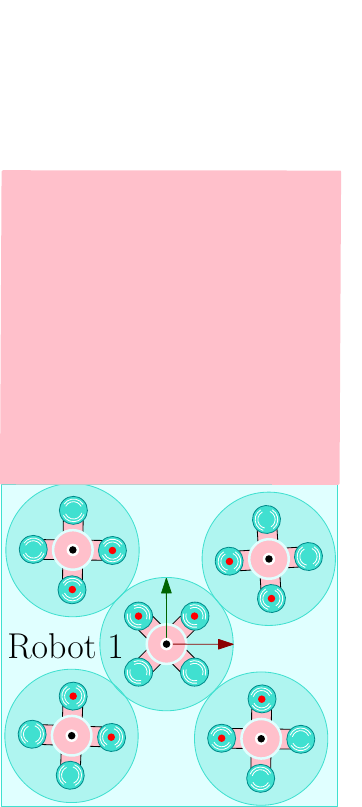}}
    \quad
    \subfloat[\label{fig:1b}]{\includegraphics[height = 6cm, trim={0cm 0.cm 0cm 0cm}, clip]{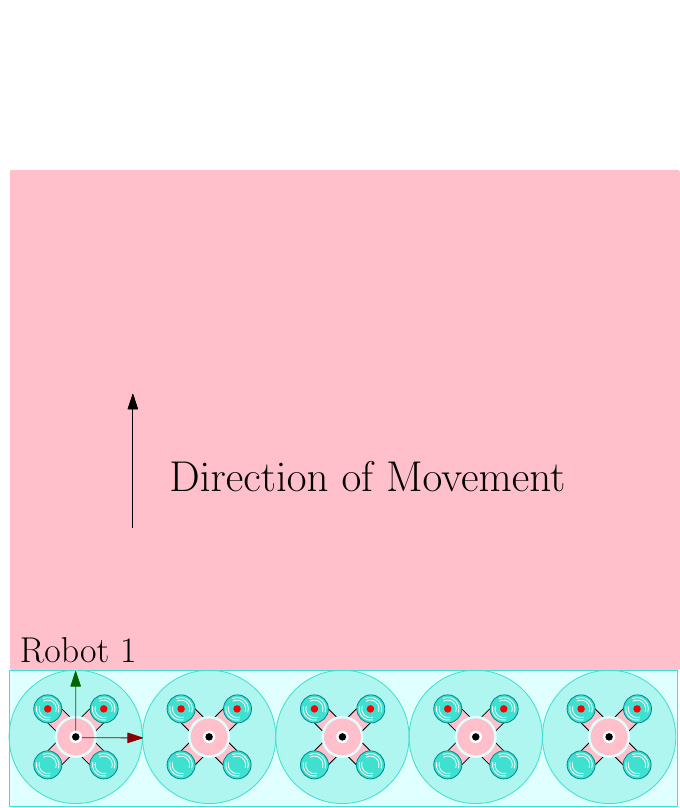}}
    \caption{Comparing the coverage span of two formations. The circles represent the camera's field-of-view of each robot, and the red dots denote the location of the ranging tags. (a) The robots are clustered together to ensure high relative pose estimation accuracy, as shown in \cite{Charles2022OptimalMF}. (b) The robots are spread apart in a horizontal line to cover a larger area, which minimizes coverage time.}
    \label{fig:formation}
    \vspace{-0.3cm}
\end{figure}

For anchor-free localization, fusing range measurements from two tags in each robot with inertial measurement unit (IMU) data using an extended Kalman filter (EKF) provides reliable relative pose estimates \cite{Shalaby2021RP,Shalaby2023MR}. This setup relaxes all motion impositions such as the robots' need to be in persistent relative motion or the need for periodic line-of-sight between the cameras and the robots. However, even with two tags per robot, there are a finite number of non-unique solutions to the relative pose estimation problem, referred to as \emph{ambiguities}. The presence of ambiguities causes the estimator to diverge in certain formations, such as when all the robots are in a straight line, as shown in Fig.~\ref{fig:1b} \cite{Shabbir2024GSF,Charles2022OptimalMF}. Despite using estimators suitable for handling these ambiguities, such as a Gaussian sum filter \cite{Shabbir2024GSF}, maintaining these formations for a long period of time may still lead to estimator divergence.

To address this issue, \cite{Charles2022OptimalMF} suggests keeping the team of robots in formations where they are close and clustered together, as shown in Fig.~\ref{fig:1a}, which theoretically maximizes the relative pose estimation accuracy for two-tagged robots. 
However, these clustered formations are not ideal for applications such as infrastructure inspection or surveillance, where maximizing coverage is beneficial. An example of robot clustering resulting in reduced coverage is shown in Fig.~\ref{fig:formation}.

This paper addresses the contrasting objectives of determining multi-robot formations that both (1) maximize coverage and (2) ensure close proximity between robots for good relative localization accuracy. Other multi-robot path planning mechanisms have focused on distributing the robots into different sectors in a large area, where each robot individually covers its sector to minimize overall coverage time \cite{Chen2021EfficientMC, Tang2023, Gao2018OptimalMC, Lee2023CA, Xiaoguang2018}. The robots generally localize themselves using the Global Positioning System (GPS). However, with a UWB ranging-based approach, the robots cannot be distributed into sectors since they must be in proximity to each other to achieve high relative pose estimation accuracy, as highlighted in \cite{Charles2022OptimalMF}.

The key contribution of this paper is a cost function that brings the robots to any desirable formation, such as a ``high-coverage'' straight-line formation, while simultaneously maintaining high relative localization accuracy. This cost function has a component that provides the user with the ability to choose the direction and distance between any two adjacent robots.
This feature enables the user to realize different formations for various applications, such as bridge inspection, as demonstrated in Section \ref{sec:bridge}. User-defined formations can be achieved using acceleration inputs \cite{Dang2019FC, Yan2021}, but the proposed component within the cost function is easily customizable and integrable with the formulation of \cite{Charles2022OptimalMF}.
Another component of this cost function allows the user to allocate a certain amount of overlap between adjacent robots' camera views, which is good for image-stitching and in improving mapping accuracy, as mentioned in \cite{Li2017}. Observability and collision avoidance terms are also incorporated into the cost function.  

The ``high-coverage'' formations generated by minimizing the proposed cost function are tested in a planning task in simulation and experiment, where the robots localize themselves and unknown anchors using a simultaneous localization and mapping (SLAM) algorithm based on the EKF. Compared to the current state-of-the-art, the proposed formations significantly reduce coverage time with minimal impact on localization accuracy.

The remainder of this paper is organized as follows. The notation and preliminaries are defined in Section \ref{sec:problem_setup}. The problem is motivated in Section \ref{sec:problem_formulation}. The proposed cost functions are in Section \ref{sec:optimization}. The application of the cost function in simulations and experiments is in Section \ref{sec:coverage}.

\section{Notation and Preliminaries}\label{sec:problem_setup}    

\begin{figure}[t]
    \vspace{0.1cm}
    \centering
    \includegraphics[width=0.79\columnwidth, trim={0cm, 0cm, 0cm, 0cm}, clip]{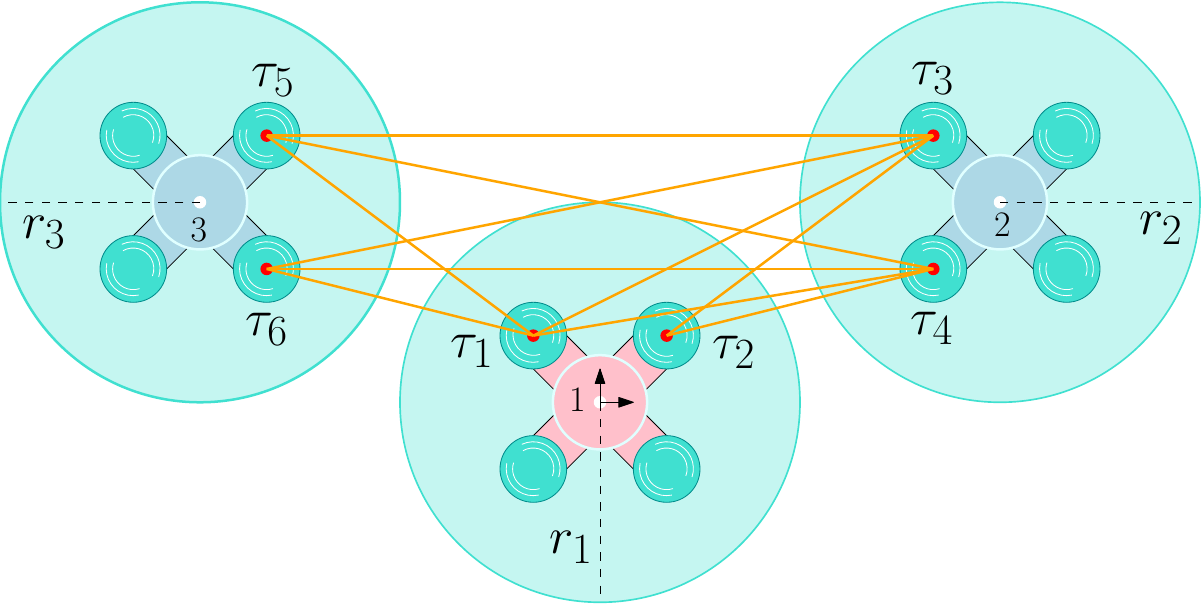}
    \caption{Problem setup for a two-tag multi-robot system, where Robot~$p$ is equipped with tags $\tau_i$ and $\tau_j$, and a camera with a circular view of radius $r_p$ in the up or down direction. Without loss of generality, the pink robot, defined as Robot~$1$, is considered to be the reference robot. 
    }
    \label{fig:setup}
    \vspace{-0.3cm}
\end{figure}

Consider $N$ robots with IDs, $\mc{P} = \{1, \ldots, N\}$. Each robot is equipped with two ranging tags, resulting in a total of $2N$ tags collectively, as shown in Fig.~\ref{fig:setup}. The physical points $\tau_1, \ldots, \tau_{2N}$ denote the location of the tags on the robots. The set of tag IDs is denoted as $\mc{V}~=~\{1, \ldots, 2N\}$. 
Each robot is assumed to be equipped with a downward or upward-facing camera that has a circular field-of-view with a known radius, $r_p$. The set of radii is denoted as $\mc{R} = \{r_1, \ldots, r_N\}$. The set $\mc{E}$ denotes the inter-tag range measurements. The bolded $\mbf{1}$ and $\mbf{0}$ are appropriately sized identity and zero matrices, respectively. Subscripts such as $\mbf{1}_{2 \times 2}$ and $\mbf{0}_{2 \times 1}$ may be used to explicitly indicate dimensions. 

A 2-dimensional orthonormal reference frame $\mc{F}_p$ is attached to Robot~$p$. A common global reference frame and a static point are denoted by $\mc{F}_g$ and $w$, respectively. 
The position of a chosen reference point in Robot~$p$ relative to point $w$, resolved in $\mc{F}_p$ is denoted $\mbf{r}^{pw}_p \in \rnums^2$. Vectors resolved in different frames are related by the transformation $\mbf{r}_p^{pw}=\mbf{C}_{pq} \mbf{r}_q^{pw}$, $\mbf{C}_{pq} \in \SO{2}$, where $\SO{2}$ is the special Orthogonal group in 2D. For conciseness, Robot~$p$ is referred to as $\text{R}_p$ in plot legends. 
The relative pose between Robots $p$ and $q$ is 
\begin{align}
    \mbf{T}_{pq} = \bbm
    \mbf{C}_{pq} & \mbf{r}^{qp}_p \\
    \mbf{0} & 1
    \ebm  \in \SE{2},
\end{align}
where $\SE{2}$ is the special Euclidean group in 2D. The exponential map of $\SE{2}$ is denoted $\exp: \se{2}~\rightarrow~\SE{2}$, where $\se{2}$ is the Lie algebra of $\SE{2}$. The ``wedge'' operator is denoted $(\cdot)^\wedge: \rnums^3 \rightarrow \se{2}$. 

The poses of all the robots are expressed relative to Robot~$1$, which is arbitrarily chosen to be the reference robot. As such, the state of the system is
\begin{align}\
    \label{eq:state_def}
  \mbf{x} = (\mbf{T}_{12}, \ldots, \mbf{T}_{1N}) \in \SE{2}^{N-1}.
\end{align}
Denoting $\delta\mbc{\xi}_p \in \rnums^{3}$, and $\mbfdel{x} = [\delta\mbc{\xi}_2^\trans \cdots \delta\mbc{\xi}_N^\trans]^\trans \in \rnums^{3\times(N-1)}$, 
the $\oplus$ operator is defined as,
\begin{align}
    \label{eq:oplus}
    \mbf{x} \oplus \delta \mbf{x} = (\mbf{T}_{12}\exp(\delta\mbc{\xi}_2^\wedge), \ldots, \mbf{T}_{1N}\exp(\delta\mbc{\xi}_N^\wedge)).
\end{align}
The position of Robot $p$ relative to Robot $q$, resolved in $\mc{F}_1$, is 
\begin{align}
    \mbf{r}^{pq}_1 = \mbf{D}\mbf{T}_{1p}\mbf{b} - \mbf{D}\mbf{T}_{1q}\mbf{b},
\end{align}
where $\mbf{D} = [\mbf{1}_{2 \times 2}\; \mbf{0}_{2\times1}]$, $\mbf{b} = [\mbf{0}_{1\times2}\;1]^\trans$.

The range measurement of Tag $i$ relative to Tag $j$ in Robots $p$ and $q$, respectively, is modelled as
\begin{align} \label{eq:meas_model}
    y_{ij} (\mbf{x}) &= \norm{ \mbf{D}\mbf{T}_{1p}
    \mbftilde{r}^{\tau_i p}_p
    - \mbf{D} \mbf{T}_{1q}
    \mbftilde{r}^{\tau_j q}_q
     } + \eta_{ij},
\end{align}
where $\mbftilde{r} = [\mbf{r}^{\trans}\;1]^\trans$, and $\eta_{ij}~\sim~\mc{N}(0, \sigma^2_{ij})$. Therefore, the augmented measurement vector is,
\begin{align}
    \label{eq:meas_model2}
    \mbf{y} &= \mbf{g}(\mbf{x}) + \mbc{\eta} =\bma{ccc} \cdots & y_{ij}(\mbf{x}) & \cdots \ema^\trans + \mbc{\eta} \in \rnums^{|\mc{E}|}, \nonumber \\
    &\hspace*{0.3cm}\forall (i,j) \in \mc{E}, \mbc{\eta} \sim \mc{N}(\mbf{0}, \mbf{R}), \; \mbf{R} = \diag(\ldots, \sigma_{ij}^2,\ldots).
\end{align}
\subsection{Optimization}
This paper finds locally optimal formations by minimizing cost functions of $\mbf{x}\in \SE{2}^{N-1}$, $J(\mbf{x})$. All such cost functions are minimized using a momentum-based gradient descent algorithm. This approach is preferred over a standard gradient descent method as it allows for faster convergence to a global or local minimum \cite{Qian1999}. The state is updated from $\mbf{x}_t$ to $\mbf{x}_{t+1}$ using a perturbation $\delta \mbf{x}_t \in \rnums^{3\times(N-1)}$ as
\begin{align}
    \delta \mbf{x}_{t} &= - \Bigl(\alpha \nabla J(\mbf{x}_t) + \beta \delta \mbf{x}_{t-1}\Bigr)^\trans,\\
    \mbf{x}_{t+1} &= \mbf{x}_t \oplus \delta \mbf{x}_t,
\end{align}
where $\nabla J(\mbf{x}_t)$ is the gradient of the cost function numerically computed using finite difference \cite{Charles2020ComplexStep}, $\alpha$ is the learning rate, and $\beta$ is the momentum parameter. Throughout the paper, the parameters $\alpha=0.001$ and $\beta=0.9$ are used. The optimization is terminated when $||\delta \mbf{x}_t|| < 10^{-4}$.

\begin{figure*}[b]
    \vspace{-0.6cm}
    \centering
    \subfloat[Straight-line formation with unsorted IDs.\label{fig:c1}]{{\includegraphics[width=0.3\textwidth, trim={7.2cm 0cm 9cm 0cm}, clip]{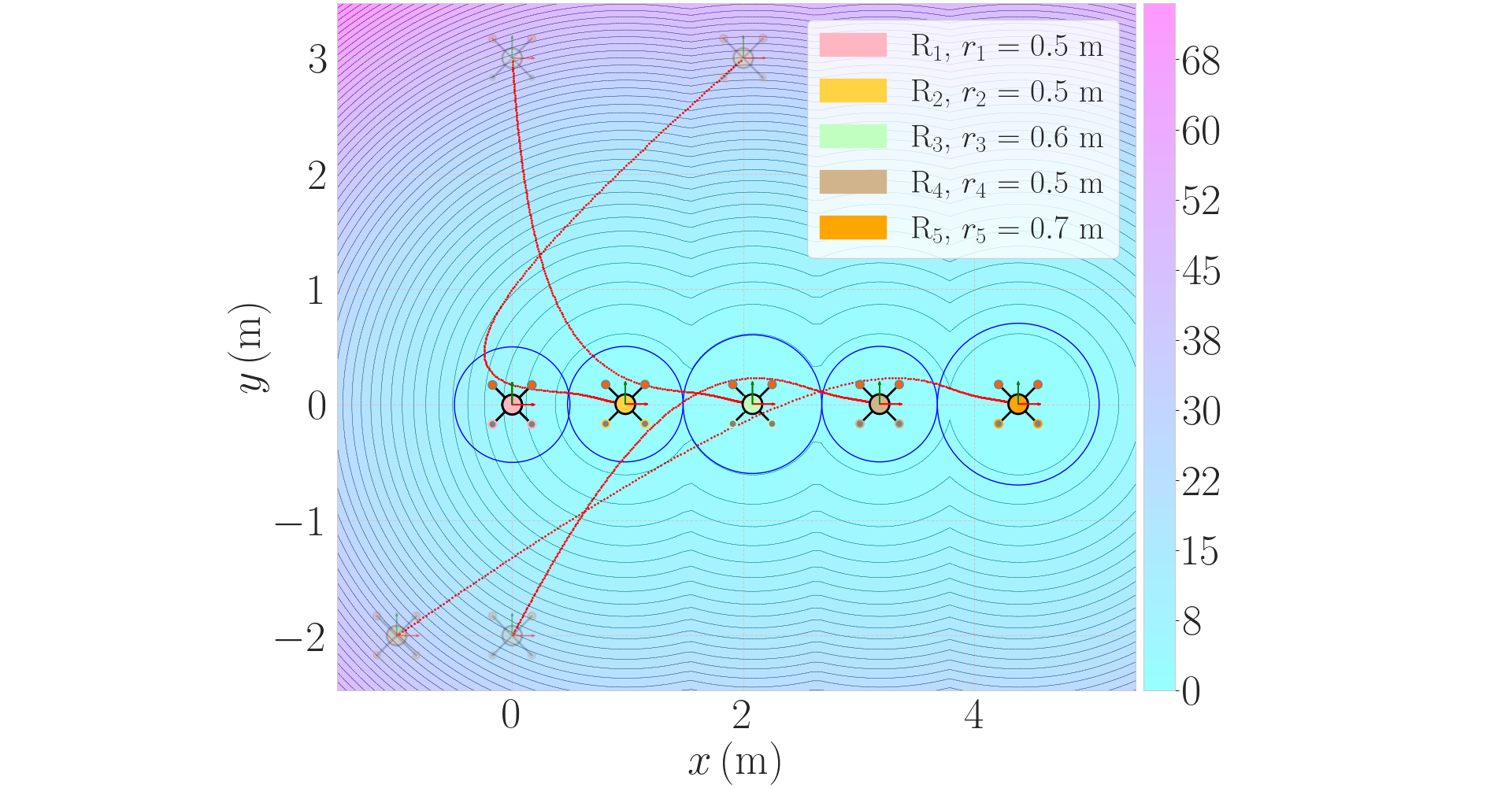}}}%
    \hspace{0.1cm}
    \subfloat[Straight-line formation with sorted IDs.\label{fig:c2}]{{\includegraphics[width=0.3\textwidth, trim={7.2cm 0cm 9cm 0cm}, clip]{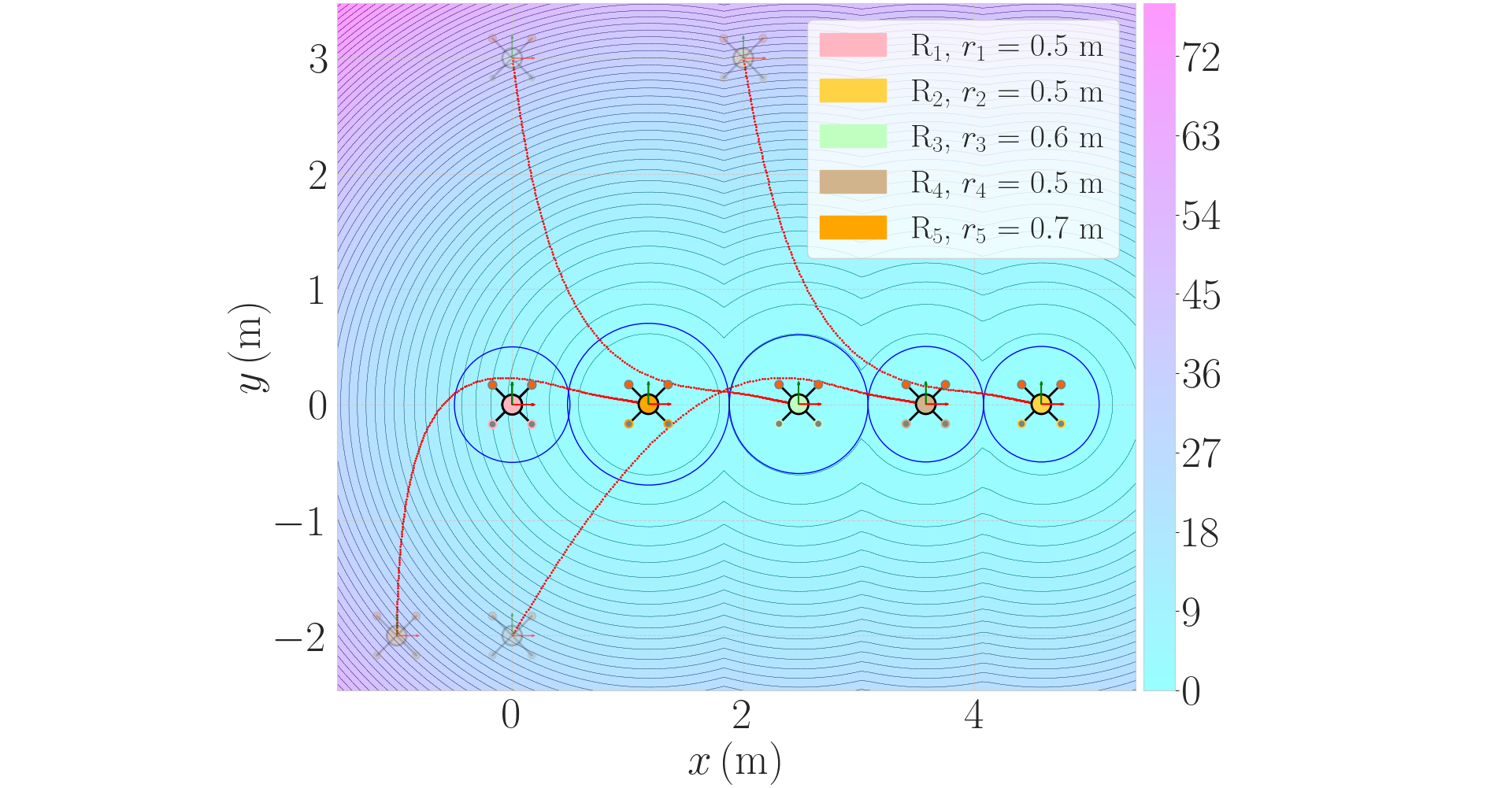} }}%
    \subfloat[V-shaped formation with sorted IDs.\label{fig:c3}]{{\includegraphics[width = 0.36\textwidth, trim={0.76cm 0cm 1.29cm 0cm}, clip]{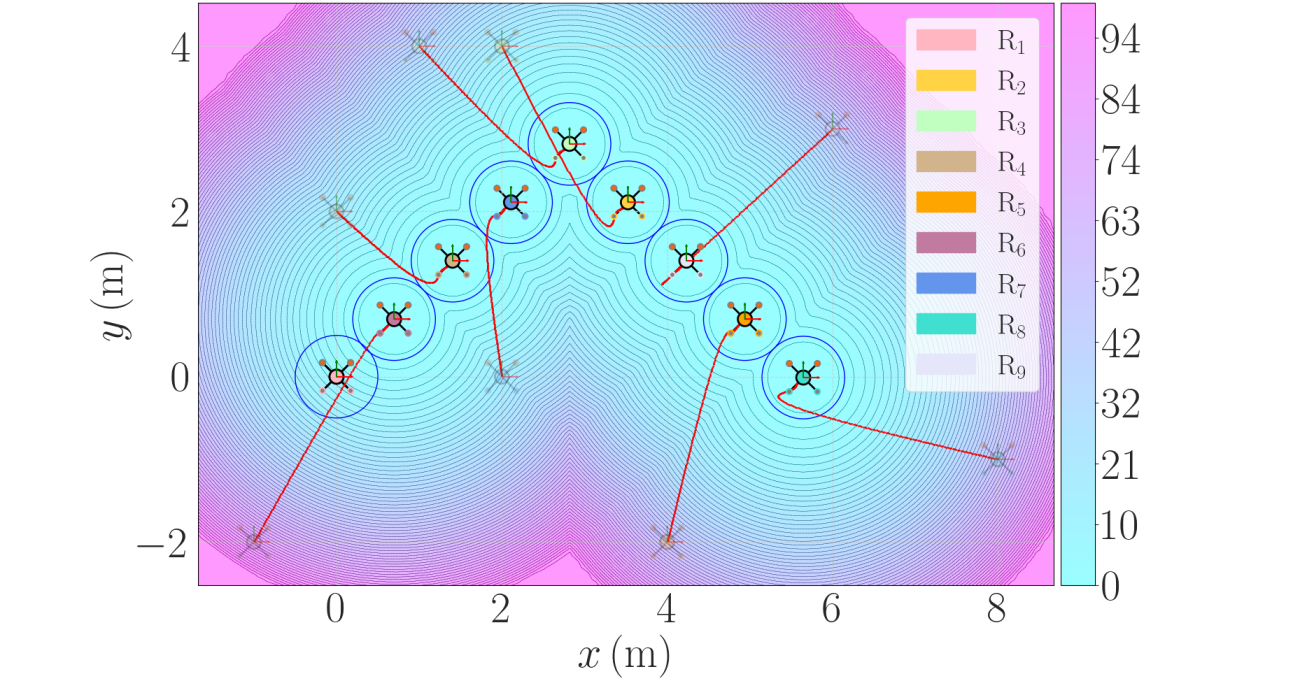}}}
    \caption{Formations obtained by minimizing $J_{\text{adj}}(\mbf{x})$. The contours represent the heatmap of the cost function $J_{\text{adj}}(\mbf{x})$, by varying the position vector, $\mbf{r}^{mn}_{n}$, between all the robots.
    }
    \label{fig:J_adj_formation}
\end{figure*}

\section{Motivation} \label{sec:problem_formulation}
The goal of this paper is to find multi-robot formations that minimize the coverage time of a given space, as shown in Fig.~\ref{fig:formation}. The challenge is to balance this objective with the necessity for accurate relative pose estimation using range measurements. To find an appropriate multi-robot formation with good ranging-based relative pose estimation accuracy, \cite{Charles2022OptimalMF} proposes the minimization of
\begin{align}
    \label{eq:opt_cost}
    J_\text{opt} (\mbf{x}) = J_\text{est} (\mbf{x}) + J_\text{col} (\mbf{x}),
\end{align}
where $J_\text{est} (\mbf{x})$ quantifies the relative pose estimation error and uncertainty using the Cramér-Rao lower bound \cite{Charles2022OptimalMF, Simon2018, Cano2023}, and $J_\text{col} (\mbf{x})$ is the collision avoidance term. Note that,
\begin{align}
    J_\text{est} (\mbf{x}) &= -\ln \det \Bigl(\mbf{H}(\mbf{x})^\trans \mbf{R}^{-1} \mbf{H}(\mbf{x})\Bigr),
\end{align}
where $\mbf{H}(\mbf{x})$ is the Jacobian of the measurement model, derived for the inter-robot range measurements in \cite{Charles2022OptimalMF}. 
The collision avoidance term is defined as \cite{Xia2016}
\begin{align}
    J^{mn}_{\text{col}}(\mbf{x}) &= \Biggl(\min\Biggl\{0, \frac{||\mbf{r}^{mn}_1||^2 - A^2}{||\mbf{r}^{mn}_1||^2 - d^2} \Biggr\}\Biggr)^2, \\
    J_{\text{col}}(\mbf{x}) &= \sum_{\substack{m, n \in \mc{P}, \\ m \neq n}} J^{mn}_{\text{col}}(\mbf{x}),
\end{align}
where $A$ is the activation radius and $d$ is the collision avoidance radius, set to $A = 0.9\,\si{m}$, and $d=0.5\,\si{m}$ throughout this paper. The multi-robot formations deduced by minimizing \eqref{eq:opt_cost} generally have the robots clustered together, where the robots have low area coverage as shown in Fig.~\ref{fig:1a}. In fact, \cite{Charles2022OptimalMF} shows that a straight-line formation with high coverage, as shown in Fig.~\ref{fig:1b}, unacceptably increases the relative pose estimation error. However, in theory, there are many ``high-coverage'' formations, possibly near the local minima of $J_{\text{est}}(\mbf{x})$, where the ranging-based relative pose estimation accuracy is high. These formations are achievable by minimizing a different cost function, as presented in Section \ref{sec:optimization}.

\section{Proposed Cost Functions} \label{sec:optimization}
Two novel cost functions are proposed in this section, which are added to \eqref{eq:opt_cost}. The first one allows any desirable multi-robot formation acquisition suitable for the task, and the second one ensures a certain degree of overlap between adjacent robots' camera views. The final cost function also takes relative localization accuracy and collision avoidance into account. Minimizing the final cost function helps the robots adopt ``high coverage'' formations, such as a ``near'' straight-line formation while ensuring consistently high accuracy in relative localization. The problem is approached in 2D since most robots, such as ground vehicles or quadcopters, only have heading as a rotational degree of freedom for planning purposes.

\subsection{Adjacent Robot Formation Cost Function}
Let $N$ robots be initially positioned at random locations. The goal of this section is to allocate the robots into any desired formation, with all formations being relative to Robot~$1$, the reference robot. The idea is to minimize the error between the actual and desired position vector between any two robots, which results in the cost function
\begin{align}
    \label{eq:adj_cost0}
    &J^{mn}_{\text{adj}}(\mbf{x}) = \Bigl|\Bigl|\mbf{r}^{mn}_{1} - \sum_{k=n}^{m-1} (r_{k+1} + r_k) \mbf{n}^{(k)}_1\Bigr|\Bigr|^2,\\
    \label{eq:adj_cost}
    &J_{\text{adj}}(\mbf{x}) = \sum_{\substack{n, m \in \mc{P}, \\ n < m}} J^{mn}_{\text{adj}}(\mbf{x}),
\end{align}
where $r_k$ and $\mbf{n}^{(k)}_1$ are user-defined parameters that determine the radial distance and direction between adjacent robots, respectively. $\mbf{n}^{(k)}_1$ is the desired unit vector associated with the position of Robot $k+1$ relative to its adjacent robot, Robot $k$, resolved in $\mc{F}_1$. All the desired unit vectors, starting with the one from the reference robot, Robot~$1$, can be written compactly as,
\begin{align}
    \mbf{n}_1 = \bma{ccc} \mbf{n}^{(1)\trans}_1 & \cdots & \mbf{n}^{(N-1)\trans}_1 \ema^\trans \in \rnums^{2\times(N-1)}.
\end{align} 
The desired position vector of Robot $m$ relative to Robot $n$, resolved in $\mc{F}_1$ is found using the summation term in \eqref{eq:adj_cost0}.

This cost function places the robots adjacent to each other in ascending order of their IDs without determining the shortest path the robots should take to form the desired formation, as shown in Fig.~\ref{fig:c1}. However, this is not ideal, and Algorithm~\ref{alg:sort} sorts the robot IDs so that the robots take the shortest path possible to the user-defined formation. This algorithm finds the permutation of the robot IDs that minimizes the overall distance traveled by the robots to reach the desired formation using the Hungarian matching algorithm \cite{Kuhn1955}, and is faster than a brute-force approach.

The sorted set of robot IDs and radii are denoted $\mc{P}_s~=~\{s_1,\ldots,s_N\}$ and $\mc{R}_s = \{r_{s_1},\ldots,r_{s_N}\}$, respectively.  For conciseness, $\mbf{r}^{s_ns_m}_{s_n}$ is denoted as $\mbfbar{r}^{nm}_n$, the attitude between robots $s_n$ and $s_m$ is denoted as $\mbfbar{C}_{nm}$, and the radius of Robot~$s_n$ is denoted as $\bar{r}_{n}$. 
For this sorted set of robot IDs, \eqref{eq:adj_cost0} becomes
\begin{align}
    \label{eq:adj_cost1}
    &J^{mn}_{\text{adj}}(\mbf{x}) = \Bigl|\Bigl|\mbfbar{r}^{mn}_1 - \sum_{k=n}^{m-1} (\bar{r}_{k+1} + \bar{r}_{k} ) \mbf{n}^{(k)}_1\Bigr|\Bigr|^2.
\end{align} 
Note that, $\mbf{n}_1$ denotes the desired unit vectors between adjacent robots starting from the reference robot, Robot~$1$, and therefore is not affected by the sorting of the IDs.

\begin{algorithm}[H]
    \caption{Sort Robot IDs by Distance To Travel}\label{alg:sort}
    Input: $\mbf{x}$, $\mc{P}$, $\mc{R}$, $\mbf{n}_1$.\\
    Output: $\mc{P}_s$, $\mc{R}_s$.
    \begin{algorithmic}[1]
    \STATE Let $\mbf{r}_1 \triangleq \bma{ccc} \mbf{r}^{21}_1 & \cdots & \mbf{r}^{N1}_1 \ema^\trans$, \\
    and $\mbf{p} = \bma{ccc} 2 & \cdots &  N\ema^\trans$, where $2, \ldots, N \in \mc{P} \setminus \{1\}$.
    \STATE $d_{\text{avg}} \leftarrow \frac{2}{N}\sum_{n=1}^N r_n$.
    \STATE Compute the approximate target locations in the goal formation, \\
    $\mbf{r}^{*}_1 \leftarrow \bma{ccc} \sum_{k=1}^2 d_{\text{avg}} \mbf{n}^{(k)\trans}_1 & \cdots  & \sum_{k=1}^N d_{\text{avg}} \mbf{n}^{(k)\trans}_1\ema^\trans$\\
    $\quad \; \triangleq \bma{ccc} \mbf{r}^{d_2d_1\trans}_1 & \cdots & \mbf{r}^{d_Nd_1\trans}_1 \ema^\trans$.
    \STATE Create a matrix cost function based on the distance traveled by each robot to the goal formation, \\
    $\mbf{C}(i,j) \leftarrow ||\mbf{r}^{*}_1(i) - \mbf{r}_1(j)||^2$ for $i,j \in \{1,\ldots,N-1\}$.
    \STATE Let $\mbf{P}$ be a permutation matrix, and $\trace(\cdot)$ is the trace operator. Find the permutation matrix that minimizes the overall distance traveled by the robots using the Hungarian matching algorithm \cite{Kuhn1955}, \quad
    $\mbf{P}^* \leftarrow \underset{\mbf{P}}{\min}\;\trace(\mbf{C}\mbf{P})$.
    \STATE $\mc{P}_s \leftarrow \{1\} \cup \{i^\text{th} \text{ element of }\mbf{P}^* \mbf{p}\} \triangleq \{s_1,\ldots, s_N\}$.
    \STATE $\mc{R}_s \leftarrow \{r_{s_n}\}$.
    \end{algorithmic}
\end{algorithm}

Fig.~\ref{fig:c2} depicts a straight-line formation acquisition by minimizing $J_\text{adj} (\mbf{x})$ with sorted robot IDs. With sorted IDs, the robots reach a straight-line formation by traveling a shorter overall distance compared to the one with unsorted IDs, shown in Fig.~\ref{fig:c1}. In both cases $\mbf{n}^{(k)}_1~=~[1\quad 0]^\trans,\, k~=~1,\ldots,N-1$.


Another instance of the implementation of this cost function is shown in  Fig.~\ref{fig:c3}, where the robots are in a V-shaped formation.
The parameters used for this example are $\mbf{n}^{(k)}_1 = [1\quad 1]^\trans,\, k = 1,\ldots,4$, $\mbf{n}^{(k)}_1 = [1\;-1]^\trans, \, k = 5,\ldots,8$, and radii $\bar{r}_k = 0.5\,\si{m}$. 

In the rest of this paper, unless $\mbf{n}_1$ is stated, the sorted set of IDs is computed using $\mbf{n}^{(k)}_1=[1\quad 0]^\trans, k~=~1,\ldots, N-1$, to maximize coverage span in the $x$-direction.


\subsection{Camera Overlap Cost Function}
To simultaneously enable overlap of the camera views of adjacent robots, and to ensure that no more than two adjacent camera views overlap, which in turn helps in maximizing coverage, minimizing the cost function 
\begin{align}
    \label{eq:overlap_cost}
    &J^{mn}_{\text{overlap}}(\mbf{x}) = \nonumber\\
    &\Bigl|\Bigl|\mbfbar{r}^{mn}_{1} - (1-\lambda)\Bigl(2\sum_{k=n}^m \bar{r}_{k} - \bar{r}_{n} - \bar{r}_{m} \Bigr) \mbfbar{n}^{mn}_1\Bigr|\Bigr|^2,\\
    &J_{\text{overlap}}(\mbf{x}) = \sum_{\substack{s_n, s_m \in \mc{P}_s,\\ n < m}} J^{mn}_{\text{overlap}}(\mbf{x})
\end{align}
is proposed, where $\lambda \in [0, 1]$ represents the percentage of the radial distance between the robots that overlap. The direction vector $\mbfbar{n}^{mn}_1$ is the unit vector pointing from Robot~$s_n$ to Robot~$s_m$ in the body frame of Robot~$1$ and is given by
\begin{align}
    \mbfbar{n}^{mn}_1 = \frac{\mbfbar{r}^{mn}_{1}}{||\mbfbar{r}^{mn}_{1}||}.
\end{align}

\begin{figure}[t]
    \vspace{0.1cm}
    \centering
    \subfloat{{\includegraphics[width=0.49\columnwidth, trim={0cm 0cm 0cm 0cm}, clip]{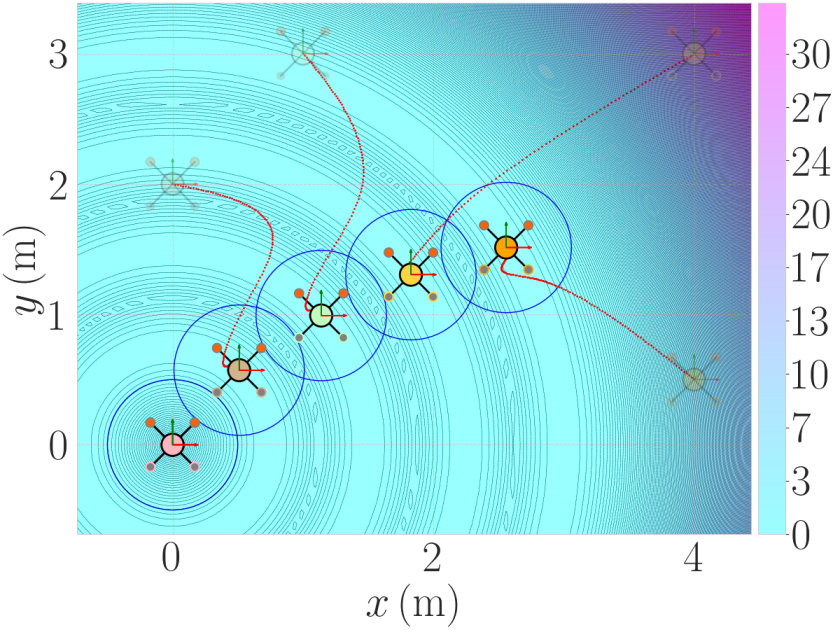}}}%
    \hspace{-0.1cm}
    \subfloat{{\includegraphics[width=0.49\columnwidth, trim={0cm 0cm 0cm 0cm}, clip]{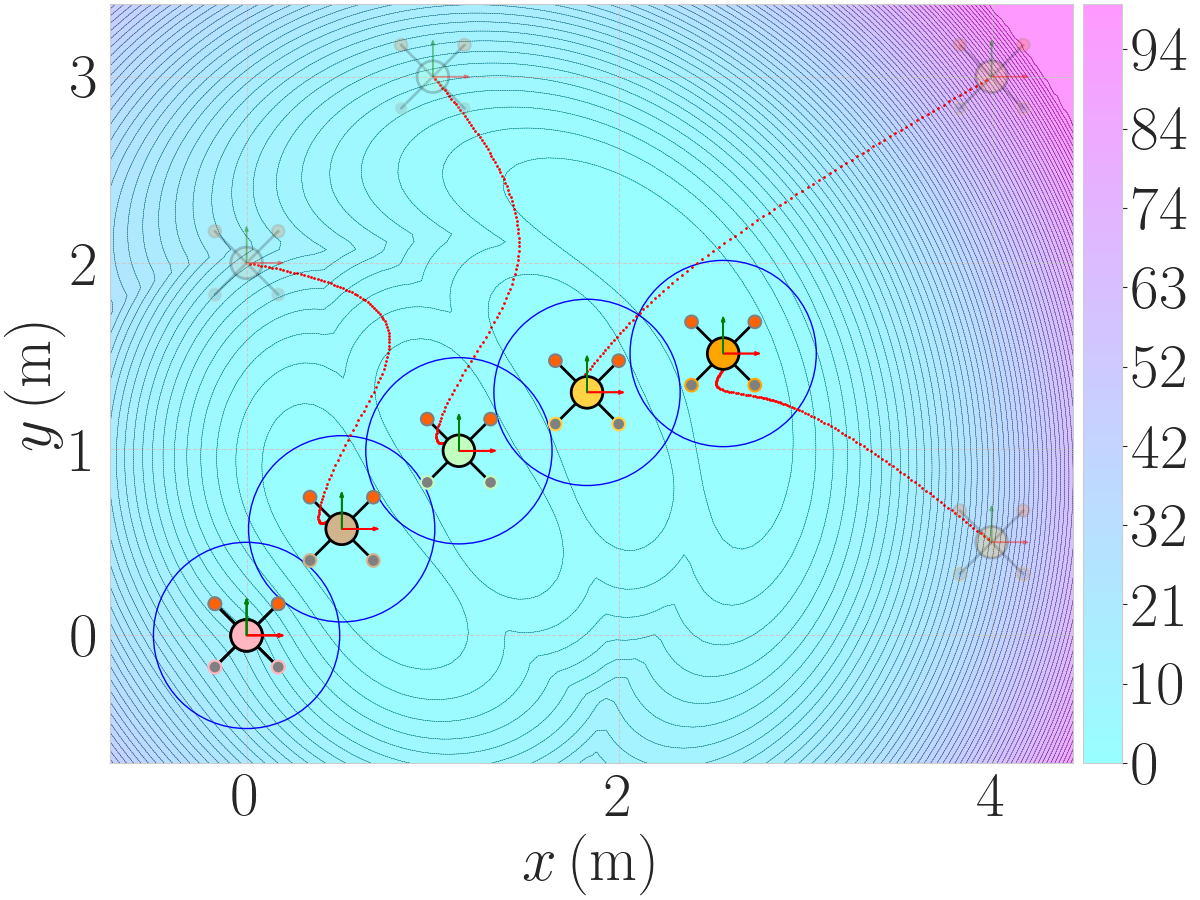} }}%
    \caption{The formation with adjacent camera overlap after minimizing $J_{\text{overlap}}$, with $\lambda = 0.25$. The left plot shows the effects of the heatmap of $J_{\text{overlap}}(\mbf{x})$ from the perspective of only Robot~$1$, and the right plot shows the effects of the heatmap from the perspective of all the robots. Only position $\mbf{r}^{mn}_{n}$ is varied between all the robots to generate the heatmaps.
    }%
    \label{fig:overlap}%
\end{figure}

An example formation with $\lambda = 0.25$ is shown in Fig.~\ref{fig:overlap}. From the contours in the left plot, note that the cost function is designed to create valleys at a distance equivalent to the summation term in \eqref{eq:overlap_cost} scaled by $(1-\lambda)$ around Robot~$1$, and similar valleys exist around all other robots. The intersection of these valleys causes the robots to overlap their camera views with adjacent robots. The advantage of this cost function is that, regardless of where the robots are initially located, every robot will end up overlapping its camera's field-of-view with adjacent robots. Therefore, this cost function is not limited to any specific formation. 


\subsection{Overall Cost Function}
\begin{figure}[t]
    \centering
    \subfloat{{\includegraphics[width=0.7\columnwidth, trim={0cm 6.8cm 0cm 8cm}, clip]{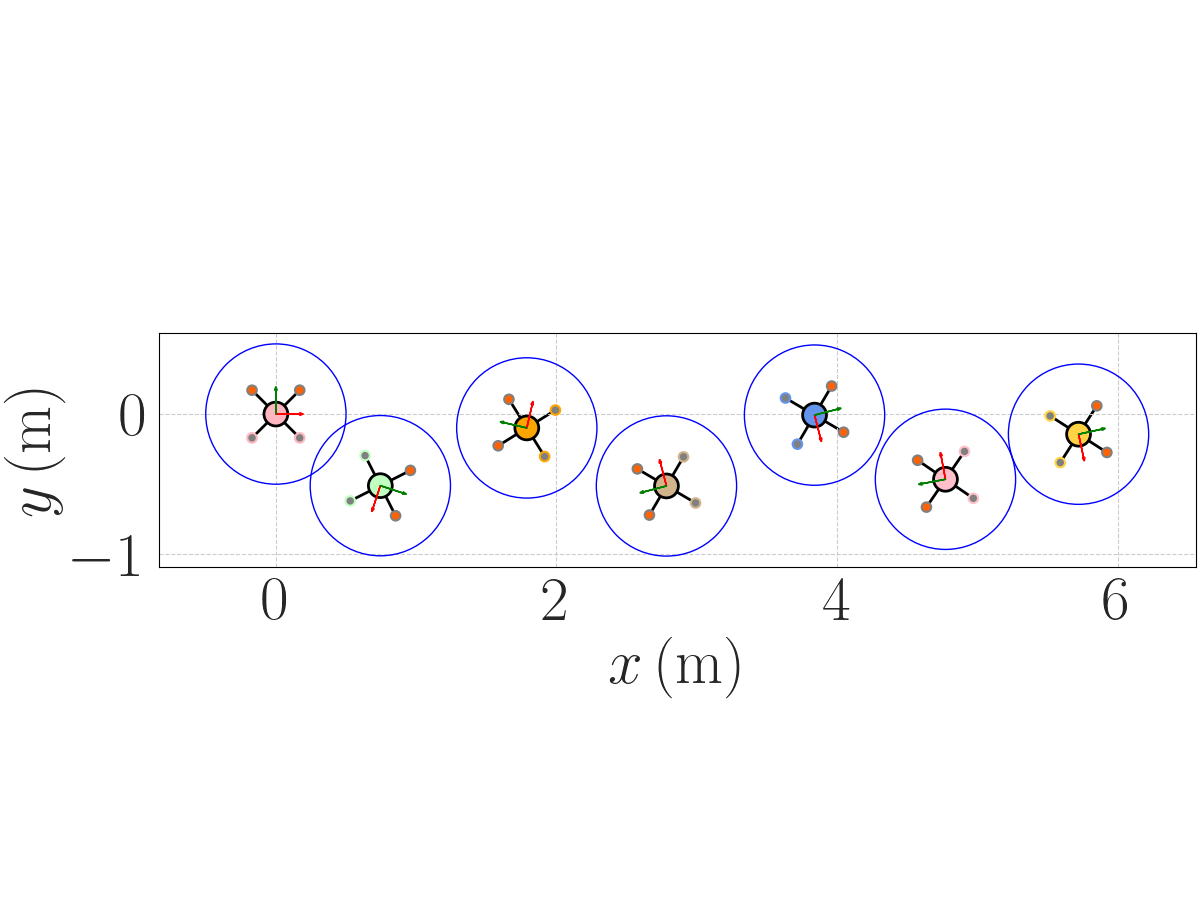} }}%
    \quad
    \subfloat{{\includegraphics[width=0.7\columnwidth, trim={0cm 5cm 0cm 8cm}, clip]{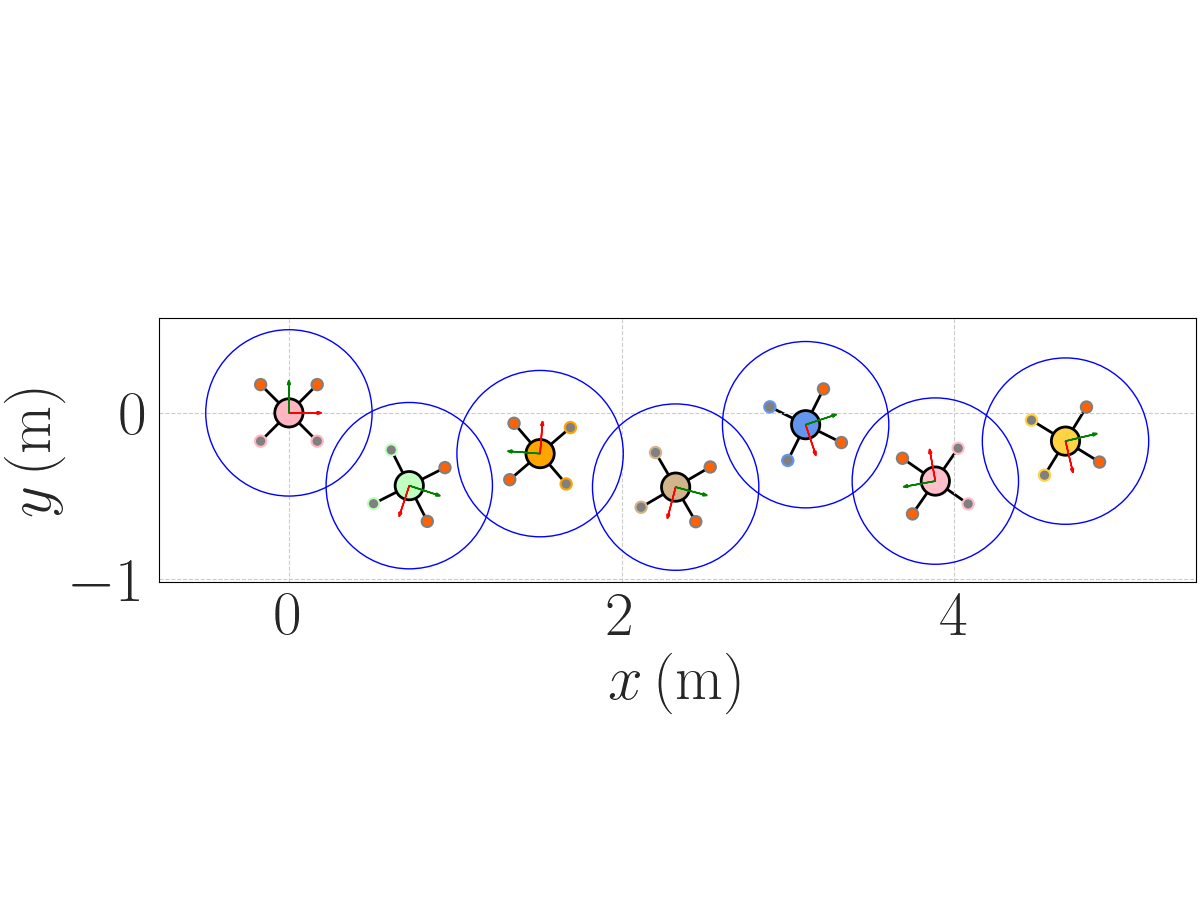} }}%
    \caption{Final formation acquisition with coverage in the $x$-direction without (top) and with (bottom) the camera overlap cost function, $J_\text{overlap}(\mbf{x})$.}%
    \vspace{-0.3cm}
    \label{fig:overlap1}%
\end{figure}

By encoding user-defined requirements for certain formations, such as a straight-line formation, and radii overlap mathematically, the proposed cost functions can be added to \eqref{eq:opt_cost} to achieve a comprehensive solution for formations that accommodate a variety of factors. These factors include the need for high coverage, the necessity for accurate relative pose estimation, and the requirement for camera overlap, among others. The overall cost function is given by,
\begin{align}
    \label{eq:opt_cost2}
    J_{\text{cov}}(\mbf{x}) = J_{\text{adj}}(\mbf{x}) + J_{\text{overlap}}(\mbf{x}) + J_{\text{est}}(\mbf{x}) + J_{\text{col}}(\mbf{x}).
\end{align}
Fig.~\ref{fig:overlap1} depicts an example formation with coverage in the $x$-direction by minimizing $J_\text{cov}(\mbf{x})$. The plots highlight the importance of $J_\text{overlap}(\mbf{x})$ in preventing the robots from non-uniformly spreading apart due to the other cost function components, notably $J_\text{adj}(\mbf{x})$. The cost $J_\text{cov}(\mbf{x})$ serves to design suitable formations for planning problems and therefore the optimization is done offline. These formation results can then be stored in the memory of the robots and used for online planning. Handling online planning initiatives like real-time non-line-of-sight issues between tags or the need for formation changes in the presence of obstacles is beyond the scope of this paper.

\subsection{Bridge Inspection Example} \label{sec:bridge}
\begin{figure}[t]  
    \centering
    \subfloat[Formation acquisition by minimizing $J_{\text{cov}}(\mbf{x})$. \label{fig:bridge_cov}]
    {
    \begin{minipage}{\columnwidth}
        \centering
    \includegraphics[width = \columnwidth, trim={0.7cm 0cm 3cm 0cm}, clip]{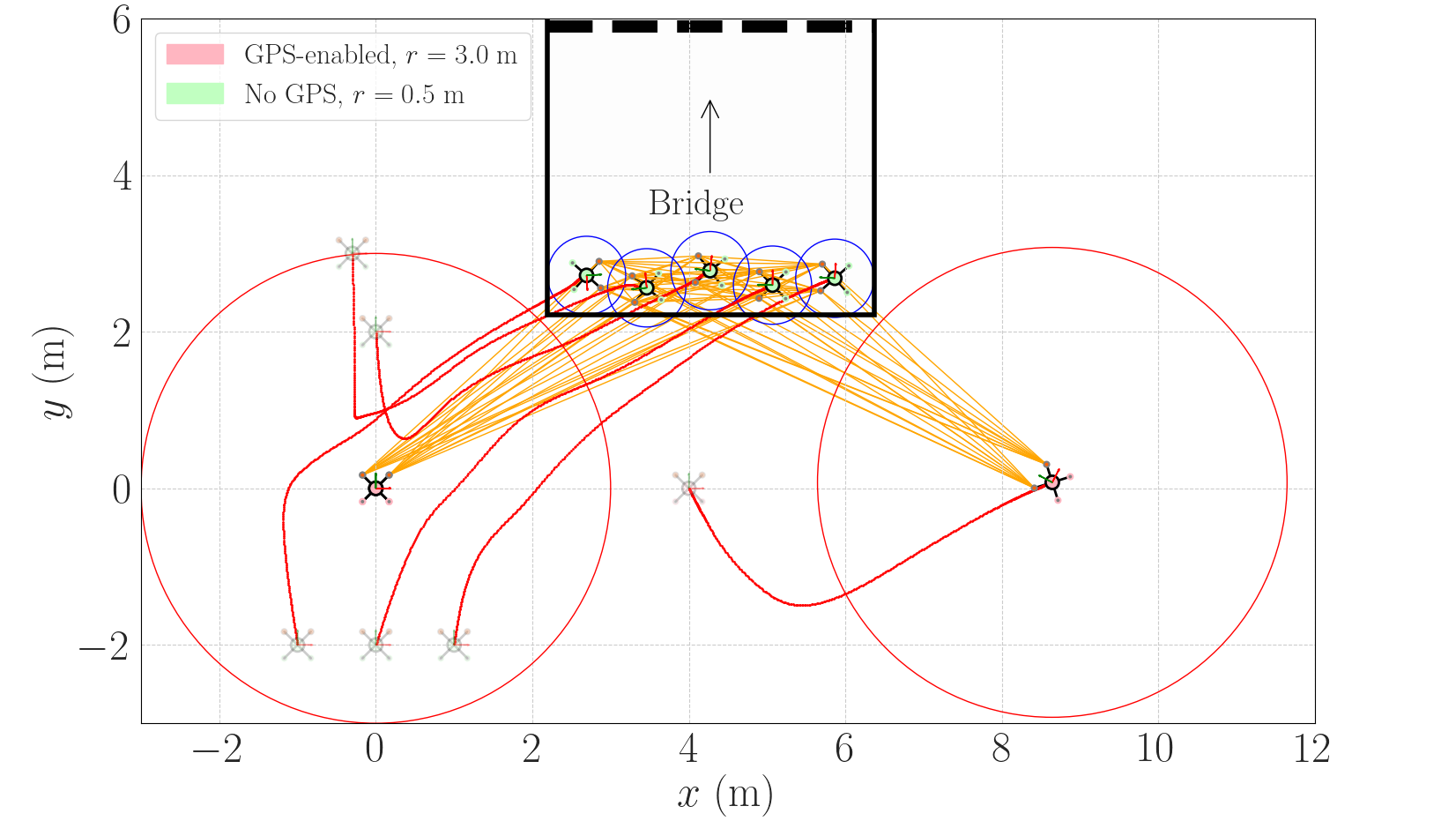} 
    \end{minipage}
    }
    \quad
    \quad
    \subfloat[Formation acquisition by randomly placing Robots $1$ and $2$ and minimizing $J_{\text{opt}}(\mbf{x})$ for the rest of the robots.\label{fig:bridge_opt}]
    {
    \begin{minipage}{\columnwidth}
    	\vspace{-0.3cm}
        \centering
    \includegraphics[width = \columnwidth, trim={0.3cm 0cm 0.3cm 0cm}, clip]{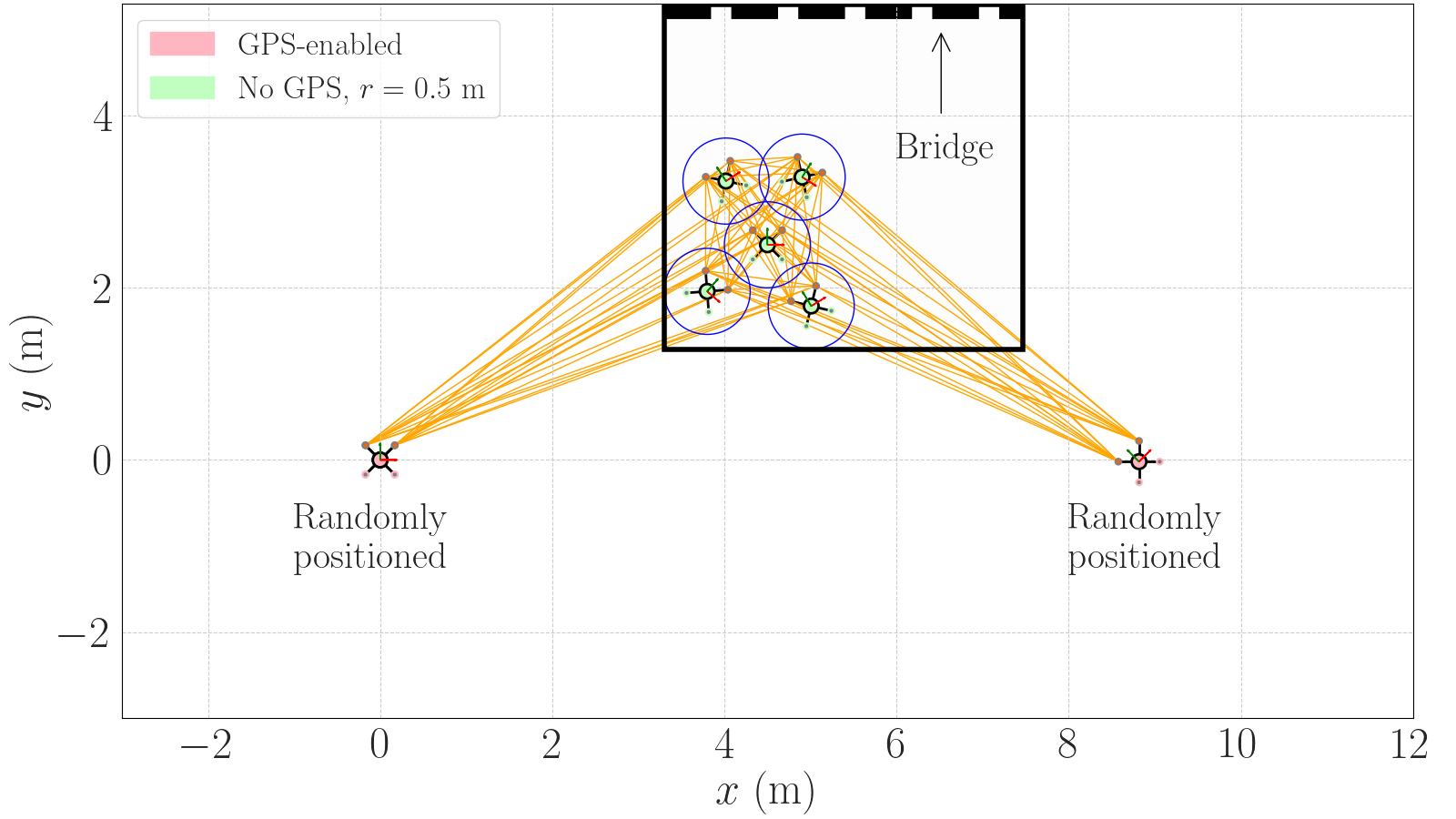} 
    \end{minipage}
    }
    \caption{Comparison of formations obtained by minimizing $J_{\text{opt}}(\mbf{x})$ and $J_{\text{cov}}(\mbf{x})$ for a bridge inspection task.}%
    \vspace{-0.3cm}
    \label{fig:bridge}
\end{figure}

The usefulness of $J_{\text{cov}}(\mbf{x})$ is shown in the bridge inspection application in Fig.~\ref{fig:bridge_cov}. Here, $5$ quadcopters with top-facing cameras inspect the underside of a bridge with no access to GPS, and two other GPS-enabled quadcopters are placed at an arbitrary angle to the inspection robots to get good localization accuracy. The desired formation is a straight-line formation of the inspection robots with some camera overlap, while ensuring that the localization accuracy is high. For $7$ robots, this is achieved by minimizing $J_{\text{cov}}(\mbf{x})$ with the parameters,
\begin{align}
    &\mbf{n}^{(1)}_1 = \bbm 1 \\ 1\ebm, \mbf{n}^{(6)}_1 = \bbm 1 \\ -1\ebm, \mbf{n}^{(k)}_1 = \bbm 1\\ 0\ebm, \quad k = 2, \ldots, 5, \nonumber \\
    \label{eq:bridge}
    &J_{\text{overlap}}^{mk}(\mbf{x}) = 0, \forall k \in \mc{P}_s\setminus\{m\}, m \in \{1,N\},
\end{align}
and there are no inter-tag range measurements between the two GPS-enabled robots. Notice that, the robots under the bridge have a ``near'' straight line formation, such that they avoid unobservable ranging-tag configurations, and are additionally aided by the GPS-enabled quadcopters to localize themselves. These planning decisions are possible because of the flexibility in customizing $J_{\text{cov}}(\mbf{x})$. In contrast, the best formation of $5$ robots obtained by minimizing $J_{\text{opt}}(\mbf{x})$ is shown in Fig.~\ref{fig:bridge_opt}. The two GPS-enabled robots are randomly placed without the help of $J_{\text{opt}}(\mbf{x})$. The inspection robots are not in a straight line, thus increasing inspection time.




\section{Application: Multi-robot Coverage} \label{sec:coverage}
\begin{figure*}[t]
    \vspace{-0.4cm}
    \centering
    \subfloat[Three tested formations.\label{fig:f1}]{{\includegraphics[height = 7.2cm, width = 0.39\textwidth, trim={11.5cm 0cm 12cm 0cm}, clip]{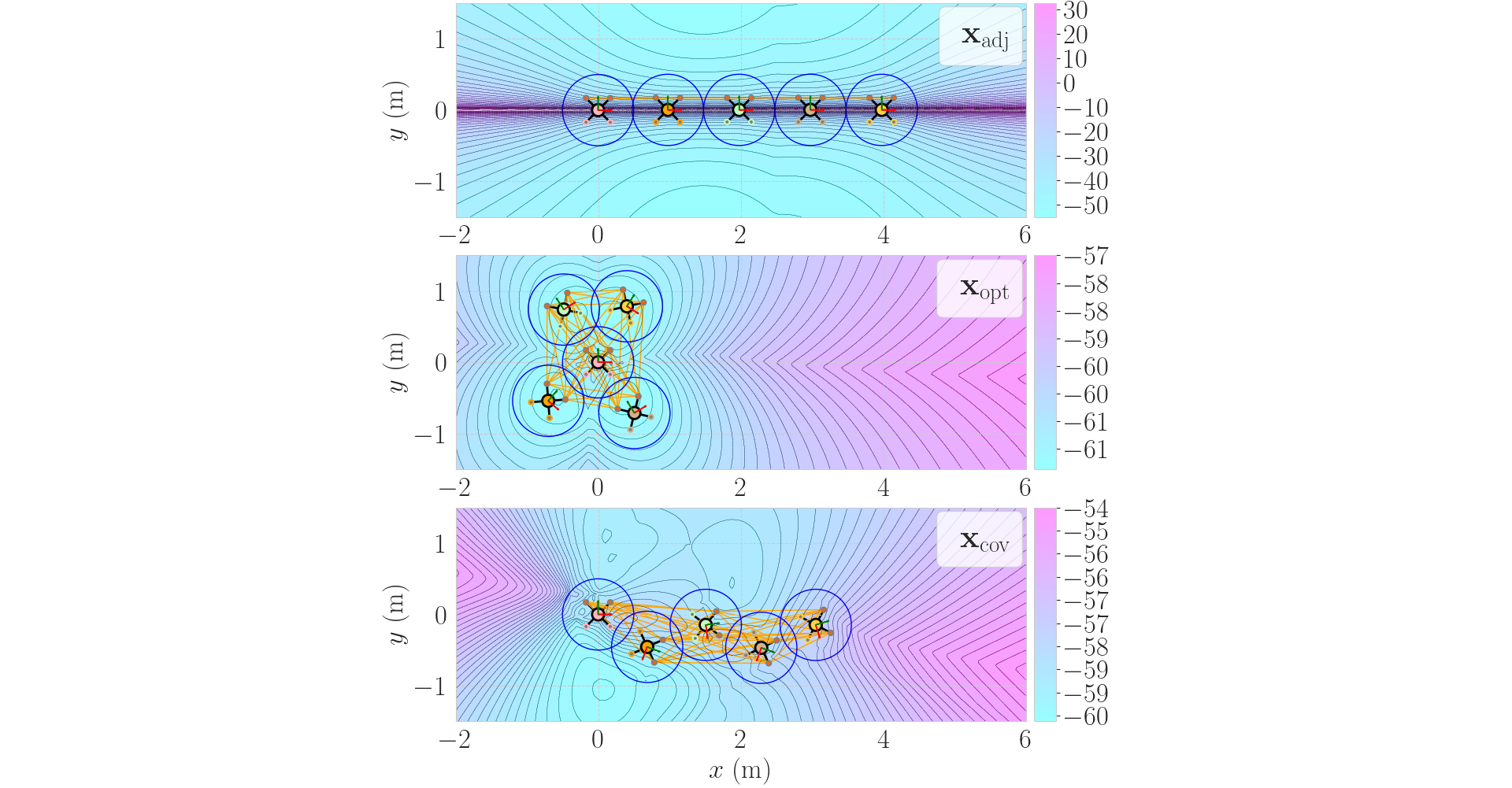}}}
    \hspace{-0.3cm}
    \subfloat[Coverage time comparison.\label{fig:f2}]{{\includegraphics[height = 7.5cm, width = 0.305\textwidth, trim={0cm 0cm 0cm 0cm}, clip]{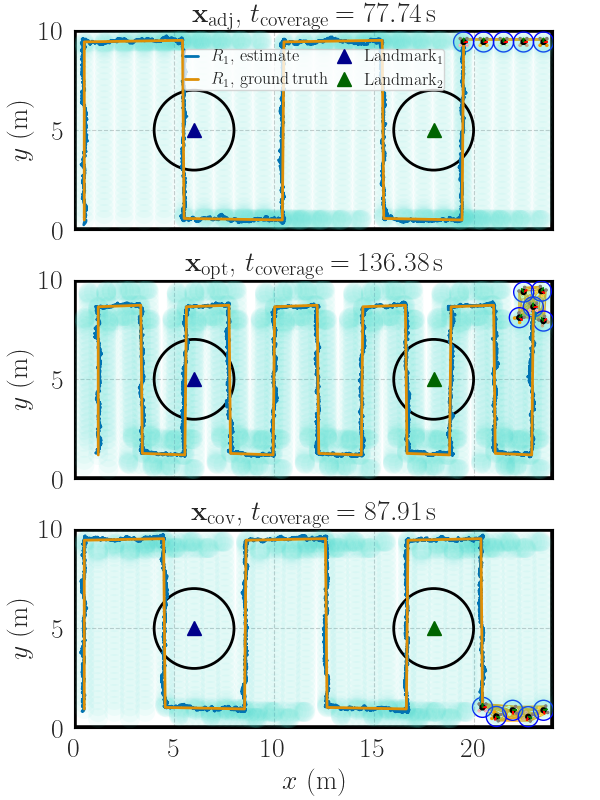}}}
    \subfloat[Estimation error comparison.\label{fig:f3}]{{\includegraphics[height = 7.5cm, width = 0.305\textwidth, trim={0cm 0cm 0cm 0cm}, clip]{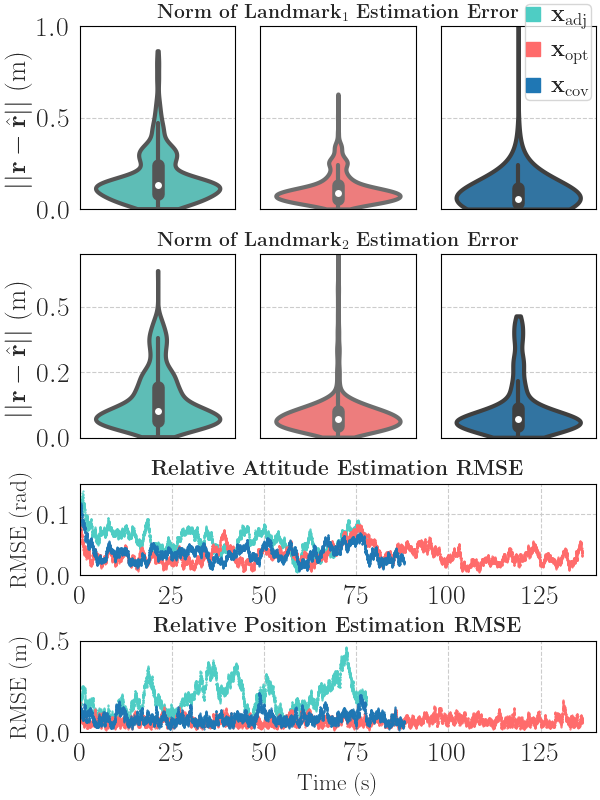}}}
    \caption{Comparison of the coverage path planning task using the three formations. (a) The heatmap of $J_\text{est}(\mbf{x})$ identifies that the straight-line formation has the highest and the cluster formation has the lowest estimation error, as expected. (b) Comparison of the coverage time for the three formations. The $\mbf{x}_{\text{cov}}$ formation has a $35.5\%$ time reduction, as compared to the $\mbf{x}_{\text{opt}}$ formation, while maintaining good relative pose estimation accuracy. (c) Various RMSE plots for the three formations over 100 Monte Carlo trials. The $\mbf{x}_{\text{cov}}$ formation has comparable inter-robot position and attitude RMSEs to the $\mbf{x}_{\text{opt}}$ formation. 
    }
    \label{fig:formations}
    \vspace{-0.3cm}
\end{figure*}

A multi-robot coverage path planning task is where the usefulness of the proposed cost function is demonstrated. The goal is to inspect a large area in a short amount of time, while ensuring good relative localization accuracy. This is achieved by minimizing $J_{\text{cov}}(\mbf{x})$ with the parameters, $\mbf{n}^{(k)}_1~=~[1\quad 0]^\trans$, $\bar{r}_k~=~0.5\, \si{m}, \,k~=~1, \ldots, N-1$, and $\lambda = 0.25$. The resultant formation is compared with a straight-line formation and a clustered formation in a coverage path planning task. These formations, along with the heatmap of $J_\text{est}(\mbf{x})$, are shown in Fig.~\ref{fig:formations}, and denoted as,
\begin{align}
    \mbf{x}_i \triangleq \argmin_{\mbf{x}} J_{i}(\mbf{x}), \quad 
    i \in \{\text{adj}, \text{opt}, \text{cov}\}.
\end{align}
The high-value regions in the heatmap of $\mbf{x}_\text{adj}$ already indicate that this formation has low relative pose estimation accuracy.
\vspace{-0.2cm}
\subsection{Simulation}
The robots are initially placed near the origin of a $10\,\si{m}~\times~24\,\si{m}$ area. They cover the space using a square-wave pattern often used in optimal coverage path planning problems \cite{Chen2021EfficientMC, Gao2018OptimalMC,Xiaoguang2018}. For simplicity, the map of the environment is assumed to be known except for the position of two static landmarks with ranging tags fitted on them. A list of waypoints is assigned to an arbitrarily chosen leader, which is Robot~$1$ here, and the other robots follow the leader in a formation using the velocity control,
\begin{align}
    \mbf{u}^{\text{reach target}/g}_n &= \mbf{u}^{\text{formation}/g}_n + \mbf{u}^{\text{waypoint}/g}_n,
\end{align}
where each control term is resolved in the robot's body frame. The components $\mbf{u}^{\text{formation}/g}_n$ and $\mbf{u}^{\text{waypoint}/g}_n$ are given in \cite[Chap. 2]{Queiroz2019}. The trajectory generated using this control law is shown in Fig.~\ref{fig:f2}. Note that, each corner of the square-wave pattern is treated as a static waypoint. Once Robot~$1$ reaches one corner in formation with the other robots, it moves to the next corner. 

The EKF-SLAM algorithm, similar to \cite{sola2014}, is used to assess the relative pose estimation accuracy. This estimation directly impacts the precision of localizing the landmarks within the context of an inspection task. EKF-SLAM is used over a batch method since it is computationally less expensive and suitable for online implementation. The interoceptive measurements are the velocity inputs in the body frame of the robots at $100\,\si{Hz}$ as shown in \cite{Shabbir2024GSF}, and the exteroceptive measurements are either inter-tag or tag-landmark range measurements at $110\,\si{Hz}$ with a covariance matrix $\mbf{R}=0.1^2\mbf{1}\,\si{m}^2$. It is assumed that the robots receive range measurements from the static landmarks only when they are within a $2\,\si{m}$ radius of the landmark. Additionally, Robot~$1$ receives GPS measurements at $50\,\si{Hz}$ with a standard deviation of $0.1\,\si{m}$ in each component to help localize itself in the global reference frame $\mc{F}_g$.

\begin{table}[t]
    \caption{ }
    \newcolumntype{Y}{>{\centering\arraybackslash}X}
    \begin{tabularx}{0.95\columnwidth}{
        |c|Y|Y|}
    \hline
    \multicolumn{3}{|c|}{\parbox{0.9\columnwidth}{\vspace{0.1cm}\centering \textbf{Percentage reduction in median estimation error with respect\\ to} $\mbf{x}_\text{adj}$ \textbf{over $100$ Monte Carlo simulations.}}} \\
    \hline
    \smaller &  $\mbf{x}_\text{opt}$ (Eq.\eqref{eq:opt_cost}) & $\mbf{x}_\text{cov}$ (proposed) \\
    \hline
    \smaller Landmark$_1$ Est. Error    & \smaller $35.4$ $\%$ & \smaller $58.8$ $\%$ \\
    \hline
    \smaller Landmark$_2$ Est. Error & \smaller $29.6$ $\%$ & \smaller $31.6$ $\%$ \\
    \hline
    \smaller Inter-robot Att. RMSE    & \smaller $47.0$ $\%$ & \smaller $40.0$ $\%$ \\
    \hline
    \smaller Inter-robot Pos. RMSE    & \smaller $66.2$ $\%$ & \smaller $59.4$ $\%$ \\
    \hline
    \end{tabularx}
    \label{tab:sim}
\end{table}

The $\mbf{x}_\text{cov}$ (proposed) formation exhibits a $35.5\%$ reduction in coverage time compared to $\mbf{x}_\text{opt}$ (clustered formation), with only $17\%$ and $11\%$ loss in relative attitude and position estimation accuracy, respectively, as shown in Fig.~\ref{fig:f2} and Fig~\ref{fig:f3}. Table~\ref{tab:sim} displays the percentage reduction in median estimation errors of $\mbf{x}_\text{opt}$ and $\mbf{x}_\text{cov}$ with respect to $\mbf{x}_\text{adj}$ for $100$ Monte Carlo simulations. 
It highlights that there is a trade-off when using $\mbf{x}_\text{cov}$ vs $\mbf{x}_\text{opt}$; $\mbf{x}_\text{cov}$ (proposed) has slightly worse inter-robot attitude and position RMSEs, but either comparable or lower landmark estimation errors than $\mbf{x}_\text{opt}$, indicating $J_\text{cov}(\mbf{x})$'s effectiveness in attaining highly observable, and ``high-coverage'' formations. The median estimation errors for $\mbf{x}_\text{cov}$ (proposed) are $0.448\,\si{m}$, $0.088\,\si{m}$, $0.032\,\si{rad}$, and $0.062\,\si{m}$ for Landmark$_1$, Landmark$_2$, inter-robot attitude, and position, respectively. This affirms that the proposed cost function allows a slight decrease in relative pose estimation accuracy to gain a significant reduction in coverage time, compared to the clustered formation, $\mbf{x}_\text{opt}$.






\begin{figure}[t]
    \vspace{0.1cm}
    \centering
    \setlength{\fboxsep}{0pt}%
    \setlength{\fboxrule}{1pt}%
    \subfloat[Experiment in progress.]{\fbox{\includegraphics[width = 0.89\columnwidth, trim={0cm 1.5cm 0cm 1.5cm}, clip]{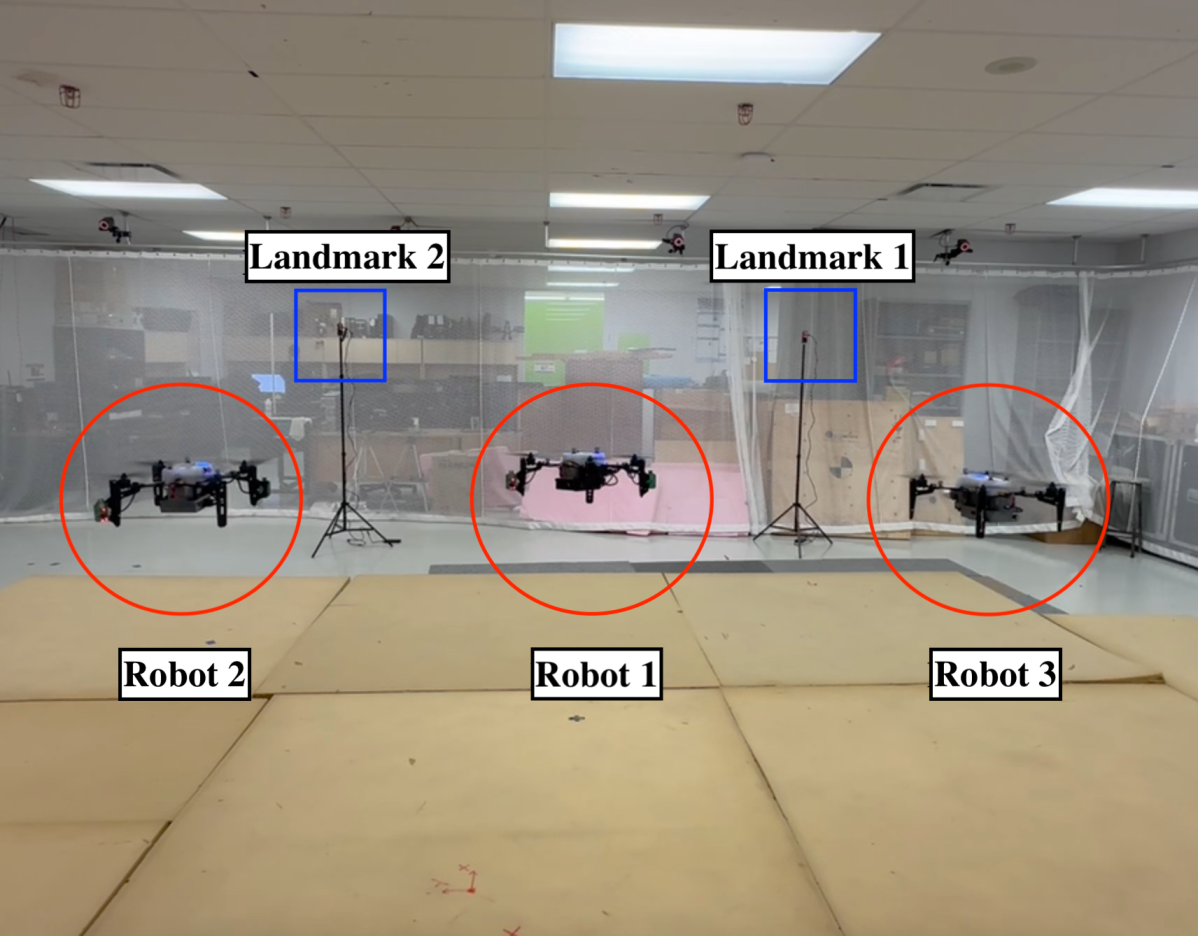}}}
    \quad 
    \subfloat[Visualization (left) and a top graphical view (right) of one of the exp-\\eriments.]{
    \begin{minipage}{\columnwidth}
        \vspace{0.0cm}
        \centering
        \includegraphics[width = 0.405\columnwidth, trim={0cm 0.cm 0cm 0cm}, clip]{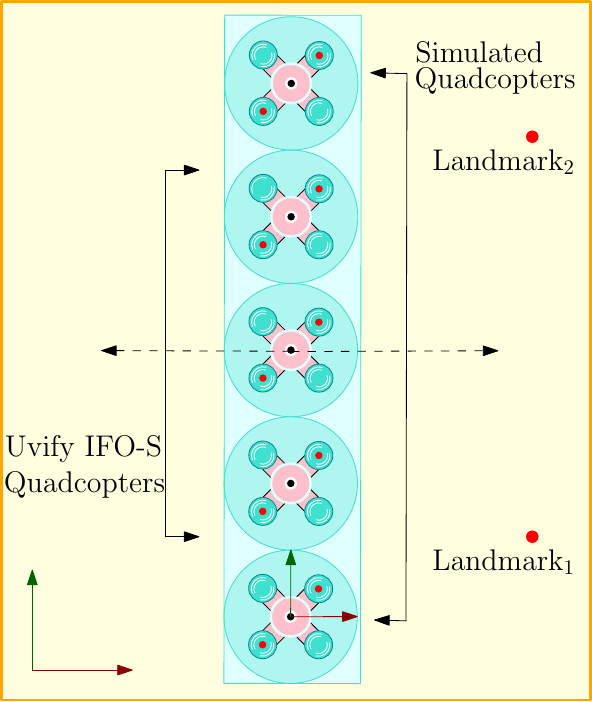}
        \includegraphics[width = 0.5\columnwidth, trim={7cm 0.cm 9.7cm 0.cm}, clip]{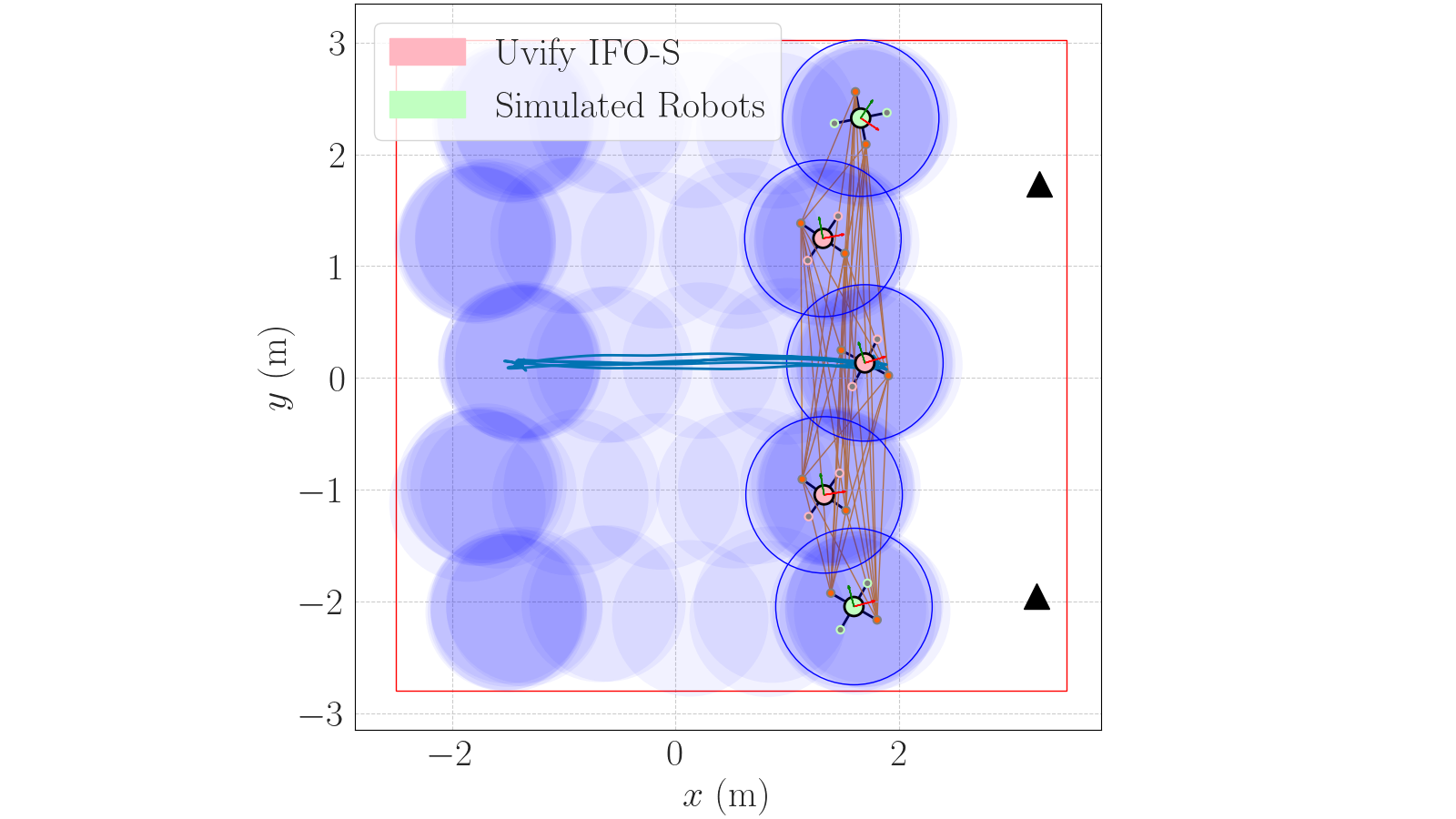}
    \end{minipage}
    }
    \caption{Experimental setup.}
    \label{fig:experiment_setup1}
    \vspace{-0.7cm}
\end{figure}

\subsection{Experiment} \label{sec:exp}
The EKF-SLAM algorithm is tested with the same formations on real quadcopters to experimentally validate that the ``high-coverage'' formations found by minimizing $J_\text{cov}(\mbf{x})$ (proposed) have good localization accuracy. Due to space limitations, each experiment is conducted with $3$ Uvify IFO-S quadcopters moving back and forth in a $4\,\si{m} \times 6\,\si{m}$ space, at a constant height, while in formation for $47\,\si{s}$. Two landmarks with UWB tags are placed at the edge of the room. The remaining two robots, with two tags each, are simulated to be in formation with the other three during the experiment. The Tags $i$ and $j$ in the robots are placed at 
\begin{align} 
    \mbf{r}^{\tau_ip}_p = \bbm 0.17 \\ -0.17 \\ -0.05\ebm, \quad
    \mbf{r}^{\tau_jp}_p = \bbm -0.17 \\ 0.17 \\ -0.05\ebm, 
\end{align}
and $r_p = 0.7,\,p \in \mc{P}$, with units in meters. Since the simulations establish that the $\mbf{x}_{\text{cov}}$ (proposed) formation reduces coverage time, the primary goal is to validate that this benefit does not significantly compromise the localization accuracy in real-world experiments. The experimental details are shown in Fig.~\ref{fig:experiment_setup1}.



The process model involves velocity inputs at $10\,\si{Hz}$ in the body frame of the robots as shown in \cite{Shabbir2024GSF}, the landmarks are static, and the measurement model involves inter-tag and tag-landmark range measurements at $80\,\si{Hz}$. For this experiment, DWM$1000$ UWB transceivers are used. The ranging protocol and UWB calibration procedure are as in \cite{Shalaby2023Calibration}. The velocity inputs with added noise are obtained by performing finite difference on ground truth position data, extracted from the Vicon motion-capture system. The added noise has a standard deviation of $0.01\,\si{rad}$ and $0.1\,\si{m}$ for the angular velocity and translational velocity components, respectively. Any interoceptive sensor data, such as IMU reading or velocity obtained using visual inertial odometry in the body frame of the robots would work as well. A covariance of $0.1^2\,\si{m}^2$ is set for the measurements received by the ranging tags in the simulated robots. Robot~$1$ is also given noisy ground truth position data as GPS measurements at $30\,\si{Hz}$ with a standard deviation of $0.1\,\si{m}$ in each component.

\begin{figure}[t]
    \centering
    \includegraphics[width = \columnwidth, trim={0cm 0.cm 0cm 0cm}, clip]{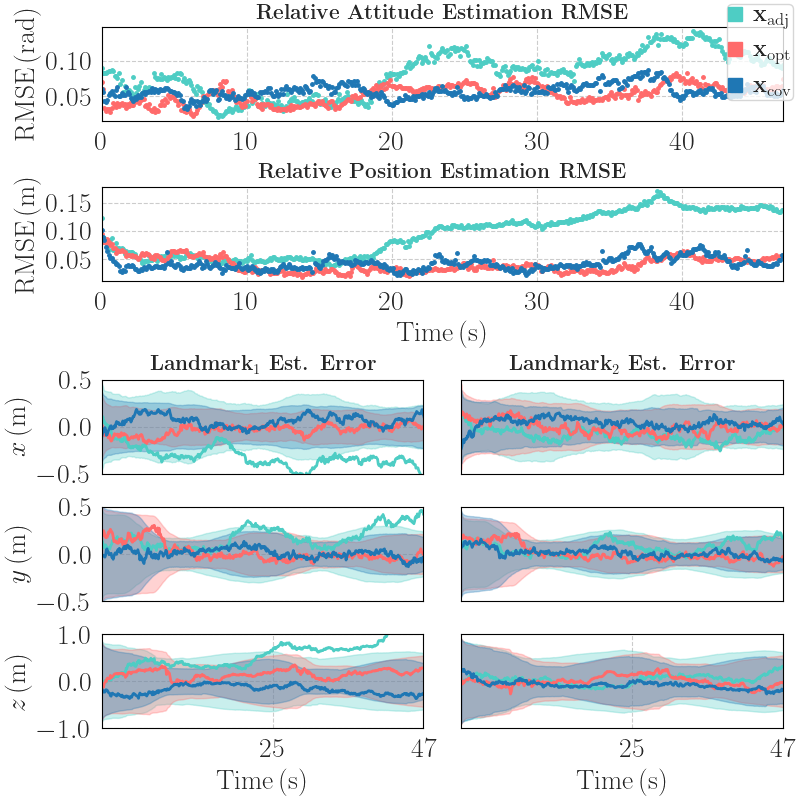}
    \caption{Different error metrics for the three formations in the experiment. The proposed formation has comparable RMSEs to the clustered formation while swiping a larger area. The shaded regions in the landmark position estimation error plots represent the $\pm 3\sigma$ bounds of the estimator.}
    \label{fig:exp_results}
    \vspace{-0.0cm}
\end{figure}

The results are shown in Fig.~\ref{fig:exp_results}. As expected, the estimator diverges for the straight-line formation due to observability issues. The landmark position and inter-robot relative pose estimation accuracy for the $\mbf{x}_{\text{cov}}$ (proposed) formation and the clustered one are similar. Furthermore, the $\mbf{x}_{\text{cov}}$ (proposed) formation maintains landmark position estimation error within the $\pm 3\sigma$ bounds, indicating low estimation error uncertainty. In Table~\ref{tab:exp}, this formation also demonstrates a significant reduction in median estimation error compared to the straight-line formation: at least $26.9\%$ for Landmark$_1$ and Landmark$_2$, and $32.9\%$ and $62.1\%$ for inter-robot attitude and position estimates, respectively, approaching levels seen in the clustered formation, $\mbf{x}_\text{opt}$. These error metrics in values are $0.112\,\si{m}$, $0.073\,\si{m}$, $0.056\,\si{rad}$, and $0.041\,\si{m}$ for Landmark$_1$, Landmark$_2$, inter-robot attitude, and position, respectively. The experiments again validate the claim of $J_\text{cov}({\mbf{x}})$ (proposed) producing ``high coverage'' formations with insignificant loss in relative pose estimation accuracy.



\begin{table}[t]
    \caption{ }
    \newcolumntype{Y}{>{\centering\arraybackslash}X}
    \begin{tabularx}{0.95\columnwidth}{
        |c|Y|Y|}
    \hline
    \multicolumn{3}{|c|}{\parbox{0.9\columnwidth}{\vspace{0.1cm}\centering \textbf{Percentage reduction in median estimation error with respect\\ to} $\mbf{x}_\text{adj}$ \textbf{for experimental data.}}} \\
    \hline
    \smaller &  $\mbf{x}_\text{opt}$ (Eq.\eqref{eq:opt_cost}) & $\mbf{x}_\text{cov}$ (proposed) \\
    \hline
    \smaller Landmark$_1$ Est. Error    & \smaller $74.1$ $\%$ & \smaller $71.1$ $\%$ \\
    \hline
    \smaller Landmark$_2$ Est. Error & \smaller $24.2$ $\%$ & \smaller $26.9$ $\%$ \\
    \hline
    \smaller Inter-robot Att. RMSE    & \smaller $32.4$ $\%$ & \smaller $32.9$ $\%$ \\
    \hline
    \smaller Inter-robot Pos. RMSE    & \smaller $64.4$ $\%$  & \smaller $62.1$ $\%$ \\
    \hline
    \end{tabularx}
    \label{tab:exp}
    \vspace{-0.35cm}
\end{table}

\section{Conclusion} \label{sec:conclusion}
This paper presents, in both simulation and experiment, that with the help of a few geometry-based constraints, ``high coverage'' formations can be achieved even if they are not optimal for inter-robot range-based relative pose estimation. The reduction in estimation accuracy for these formations is insignificant. The easy customizability of the proposed cost function to achieve ``high coverage'' formations with acceptable relative pose estimation accuracy is one of its strongest points. It can be used for a variety of applications such as multi-robot coverage, multi-robot search and rescue, and multi-robot inspection. Future work includes adopting this cost function for problems in 3D and extending the implementation of this cost function in online planning initiatives where the robots are tasked to cover a large area while avoiding obstacles.
\newpage
{\AtNextBibliography{\smaller}
\printbibliography}
\end{document}